\documentclass[runningheads]{llncs}
\usepackage{graphicx}
\usepackage{amsmath,amssymb} 
\usepackage{color}
\usepackage{epsfig}
\usepackage{subcaption}

\usepackage[breaklinks=true,bookmarks=false]{hyperref}
\usepackage[width=122mm,left=12mm,paperwidth=146mm,height=193mm,top=12mm,paperheight=217mm]{geometry}
\begin{document}
\pagestyle{headings}
\mainmatter
\def\ECCV18SubNumber{340}  

\title{Enlarging Context with Low Cost: Efficient Arithmetic Coding with Trimmed Convolution} 

\titlerunning{ECCV-18 submission ID \ECCV18SubNumber}

\authorrunning{ECCV-18 submission ID \ECCV18SubNumber}

\author{Mu Li, Shuhang Gu, David Zhang\\
Department of Computing, Hong Kong Polytechnic University\\
{\tt\small csmuli@comp.polyu.edu.hk, shuhanggu@gmail.com, csdzhang@comp.polyu.edu.hk}
\and
Wangmeng Zuo\\
School of Computer Science and Technology, Harbin Institute of Technology\\
{\tt\small cswmzuo@gmail.com}
}
\institute{Paper ID \ECCV18SubNumber}

\maketitle

\begin{abstract}
Arithmetic coding is an essential class of coding techniques which have been widely used in various data compression systems and exhibited promising performance.
 One key issue of arithmetic encoding method is to predict the probability of the current coding symbol from its context, i.e., the preceding encoded symbols, which usually can be executed by building a look-up table (LUT).
However, the complexity of LUT increases exponentially with the length of context. Thus, such solutions are limited in modeling large context, which inevitably restricts the compression performance.
Several recent deep neural network-based solutions have been developed to account for large context, but are still costly in computation.
%
%
The inefficiency of the existing methods are mainly attributed to that probability prediction is performed independently for the neighboring symbols, which actually can be efficiently conducted by shared computation.
To this end, we propose a trimmed convolutional network for arithmetic encoding (TCAE) to model large context while maintaining computational efficiency.
As for trimmed convolution, the convolutional kernels are specially trimmed to respect the compression order and context dependency of the input symbols.
Benefited from trimmed convolution, the probability prediction of all symbols can be efficiently performed in one single forward pass via a fully convolutional network.
Furthermore, to speed up the decoding process, a slope TCAE model is presented to divide the codes from a 3D code map into several blocks and remove the dependency between the codes inner one block for parallel decoding, which can $60\times$ speed up the decoding process.
Experiments show that our TCAE and slope TCAE attain better compression ratio in lossless gray image compression, and can be adopted in CNN-based lossy image compression to achieve state-of-the-art rate-distortion performance with real time encoding speed.
\end{abstract}

\section{Introduction}
Image compression aims to encode image with less bits, and can provide an effective solution to save storage requirement and transmission bandwidth.
Based on whether distortion is allowed in the reconstructed image or not, image compression can be classified into lossy and lossless methods.
For both the two categories of approaches, lossless entropy encoding is an important building block for generating the compressed representation.




One key issue of entropy encoding is to predict the probability of the current symbol to be encoded.
In some earlier approaches, such as JPEG, the frequency of the symbols is directly counted and the classical Huffman coding approach is used to compress the codes.
While, recent approaches tend to take benefit from the image context information for better predicting the probability of symbols.
In JPEG 2000, the EBCOT coder~\cite{medouakh2011entropy} is employed to model the context and approximate the probability, and the Binary Arithmetic Coding-MQ-Coder is utilized to compress the codes.
%
In H.264/AVC, the context-adaptive binary arithmetic coding (CABAC)~\cite{marpe2003context} is introduced to model the context of the preceding two encoded symbols for compressing the codes with arithmetic coding.
CABAC adopts look-up table (LUT) to predict the probability of the current symbol based on its context information.
%

The PixelCNN and PixelRNN models have validated the effectiveness of deep neural networks (DNNs) for capturing highly complex and long-range dependence between pixels.
However, since the un-decoded bits can not be utilized for probability prediction in the decoding process, standard DNNs can not be directly applied for entropy coding in image compression.    
Toderici et al.~\cite{toderici2015variable} present a recurrent neural network (RNN) based model, where a binary RNN is trained to estimate the probability of bits for better arithmetic coding.
However, their model needs to conduct one forward pass to predict each bit, making the RNN based encoder~\cite{toderici2015variable} computationally  expensive.
Recently, Li et al.~\cite{li2017learning} replace the unavailable values in the context cuboid with pre-defined values and train a shallow CNN to predict the probability of current bit.
Even though the CNN solution~\cite{li2017learning} speeds up the RNN based encoder, it also performs probability prediction independently for the neighboring symbols and remains inefficient.



\begin{figure*}[!tbp]
\begin{center}
\includegraphics[width=1.0\linewidth]{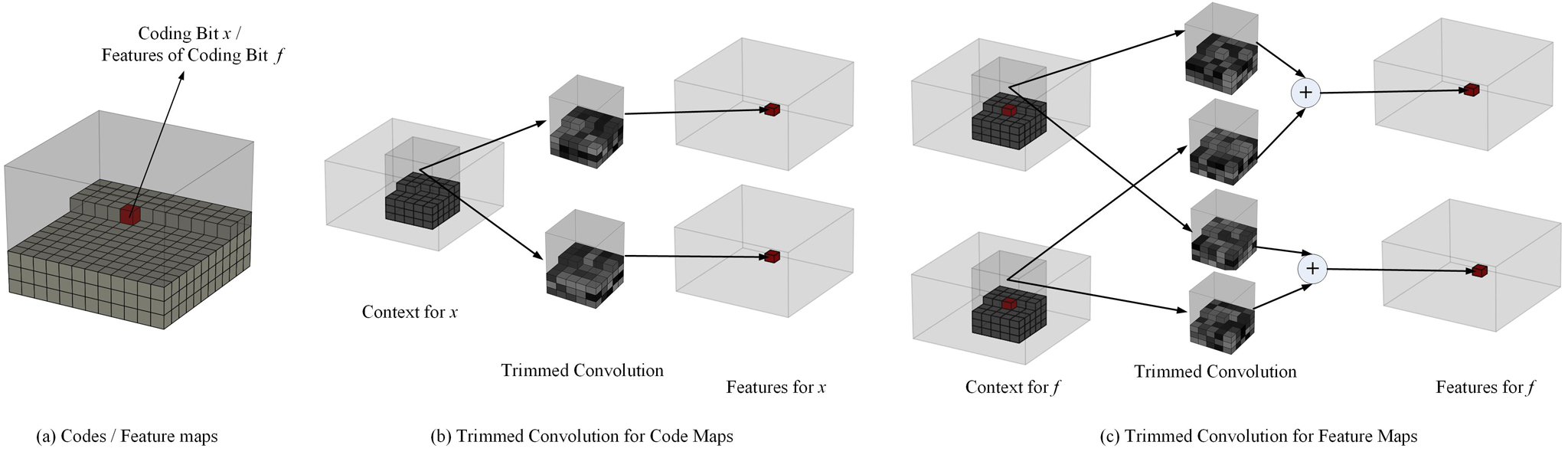}
\end{center}
   \caption{Illustration of the trimmed convolution operations used in the input layer and the hidden layers. 
   The red blocks represent the current code or feature to be encoded. 
   The blank area in the code maps or feature maps represent the non-encoded codes or features along with the encoding order.}\label{fig:simple}
\label{outline}
\end{figure*}

In this paper, we propose a trimmed convolutional network for arithmetic encoding (TCAE) which can greatly speed up the process of arithmetic encoding.
Specifically, TCAE introduces a new class of trimmed convolution operations to accelerate the probability prediction step in the entropy coding process.
Compared with the standard convolution operations which incorporate all the surrounding information to generate the output, trimmed convolution utilize a binary mask to avoid the adopting of certain input values.
Equipped with trimmed convolution kernels, the proposed TCAE approach is able to avoid the involvement of non-encoded symbols for probability prediction.
Consequently, a fully convolutional network architecture can be directly adopted for probability prediction, making our TCAE highly efficient for arithmetic encoding.





In Fig.~\ref{fig:simple} (a) and (c), we illustrate the application of trimmed convolution kernels on the the input layer and hidden feature maps, respectively. 
For the current bit to be encoded in the input layer (red pixel in Fig.~\ref{fig:simple} (a)), it should be excluded from the context for probability prediction.
While, for the value in the same location of hidden feature map, as it only conveys the information from the corresponding context area, we should include it in the coding process.
We will give more details on mask settings for sophisticated multiple filters and 3D data in Section~\ref{sec:method}.


%
Furthermore, to accelerate the decoding speed, we introduce a slope TCAE schedule, which divides the codes of 3D code maps into several blocks by requiring that the codes inner one block are independent. 
Therefore, we can decode the codes in one block simultaneously, while the code blocks still need to be decoded in order. Concretely, we set $(i,j,k)$ is the coordinate of the 3D code map $\mathbf{x}$. The $t$-th code block is defined as $CB_t(\mathbf{x})=\{x_{i,j,k}|i+j+k=t\}$. 

%
In this paper, we evaluate our TCAE and slope TCAE on both the image lossless and lossy compression tasks.
For the lossless compression task, we adopt the proposed algorithms to compress the gray images, and achieve higher compression ratio than the existing lossless compression methods, such as PNG, JPEG-LS and JPEG2000-LS.
While, for the lossy compression task, TCAE and slope TCAE is adopted to compress the intermediate codes of the lossy compression system.
We take a recently proposed CNN-based system~\cite{li2017learning} as an example, and utilize our methods to compress the binary codes and importance map generated by the encoder.
%
%
Compared with~\cite{li2017learning}, our TCAE can not only improve the compression performance due to the consideration of large context, but also be significantly faster in terms of encoding speed.
Benefited from trimmed convolution, the compression system~\cite{li2017learning} with slope TCAE can encode the image in real time, and $60\times$ speed up the decoding process.

%
%
%
To sum up, the contribution of this work is four-fold:
\begin{itemize}
  \item Trimmed convolution is incorporated with fully convolutional network to perform probability prediction to all bits within one single forward pass.
  \item A trimmed convolutional arithmetic encoder is developed to encode either gray image or intermediate codes of CNN-based lossy compression system.
  \item A slope TACE is introduced to break down the dependency of a block of codes for parallel decoding which $60\times$ speeds up the decoding process.
  \item Experiments show that for lossless gray image compression TCAE can achieve higher compression ratio than PNG and JPEG2000-LS.
        Based on the CNN-based lossy image compression system~\cite{li2017learning}, TCAE can achieve real time encoding speed without sacrifice of the rate-distortion performance.
\end{itemize}


\section{Related Work}\label{sec:related}

\subsection{Lossless image compression standards}

Lossless image compression has been investigated for decades to compress image into a smaller size without losing any information.
Most existing methods share similar pipeline, i.e., building a statistical model and then using it to guide the mapping of input data stream to bit sequences by encoding the high frequency symbol with fewer bits than low frequency symbol.
JPEG-LS~\cite{weinberger2000loco} takes use of the LOCO-I algorithm which adopts the median edge detection predictor to predict the image and then compress the residual with the context-based entropy coding.
JPEG2000-LS, the lossless version of JPEG 2000 standard~\cite{skodras2001jpeg}, is based on reversible integer wavelet transform (biorthogonal 3/5).
PNG~\cite{boutell1997png} first uses a lossless filter to transform the images and then compresses the transformed data with the DEFLATE algorithm which is a combination of LZ77 and Huffman coding.
Both TIFF and GIF apply Lemple-Ziv-Welch (LZW) algorithm~\cite{welch1984technique} which is a dictionary based method for lossless image compression.

\subsection{Entropy encoding}

Entropy encoding is a lossless data compression scheme which has played an essential role in both lossy and lossless image compression systems.
Run length encoding, Huffman coding~\cite{huffman1952method}, Golomb-Rice coding~\cite{gallager1975optimal} and arithmetic coding~\cite{witten1987arithmetic} are several representative entropy encoding techniques.
Run length coding is a very simple lossless data compression scheme, where a consecutive symbol sequence are stored as the symbol and its count.

Huffman coding~\cite{huffman1952method} is a variable-length coding scheme that encodes high frequency symbols with short codeword and low frequency symbols with long codeword.
In Golomb-Rice code~\cite{gallager1975optimal}, the input value is first divided into two parts, i.e., quotient and remainder, and truncated binary encoding is then adopted to encode the remainder.
Arithmetic coding~\cite{witten1987arithmetic} first predicts the probability of the current symbol to be encoded and then uses it to divide the current interval into sub-intervals for encoding the updated sequence.

\subsection{Deep networks based entropy encoding}
PixelCNN~\cite{van2016conditional} and PixelRNN~\cite{oord2016pixel} have shown great power in image generation and modeling highly complex and long-range dependence. However, both of them are designed for 2D image and only model the context on the 2D plane instead of 3D code maps.
Toderici et al.~\cite{toderici2016full} learn a binary recurrent network to model the context both from the same coding plane and the codes from the previous coding planes.
However, the probability of each bit should be estimated with one forward pass, making binary RNN estimator computational very inefficient.
Li et al.~\cite{li2017learning} introduce a convolutional entropy estimator to model the code context for the 3D code maps by extracting a specially defined context cube and processing the cube with a small convolutional network. 
Compared with binary RNN estimator, the convolutional estimator is relatively more efficient, but repeating calculation remains inevitable due to the overlap of the context of overbearing bits.
Actually, the inefficiency of the methods in~\cite{toderici2016full,li2017learning} is mainly attributed to that probability prediction is performed independently for the neighboring symbols.
In this work, we introduce a class of trimmed convolution to make the probability prediction more efficient by sharing computation.
By stacking multiple trimmed convolution layers, our TCAE can also exploit large context to improve compression performance.

\begin{figure}[!tbp]
\centering
\begin{minipage}{1.0\textwidth}
  \centering
\includegraphics[width=1.0\textwidth]{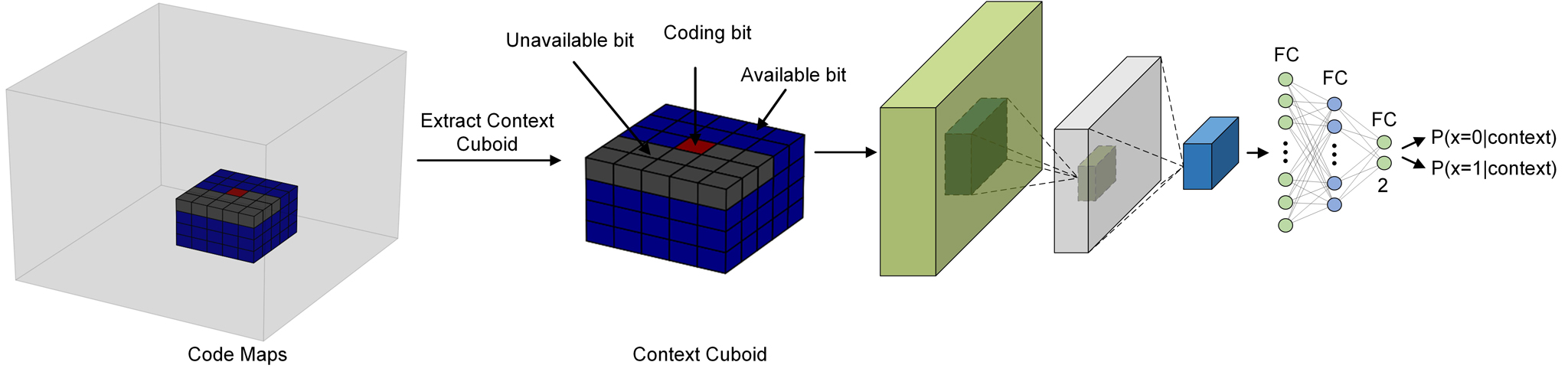}%
\subcaption{Framework for traditional convolutional network for entropy probability estimator used in~\cite{li2017learning}. One bit is processed as a sample.}\label{fig:1a}
\end{minipage}\par\medskip
\begin{minipage}{1.0\textwidth}
  \centering
\includegraphics[width=1.0\textwidth]{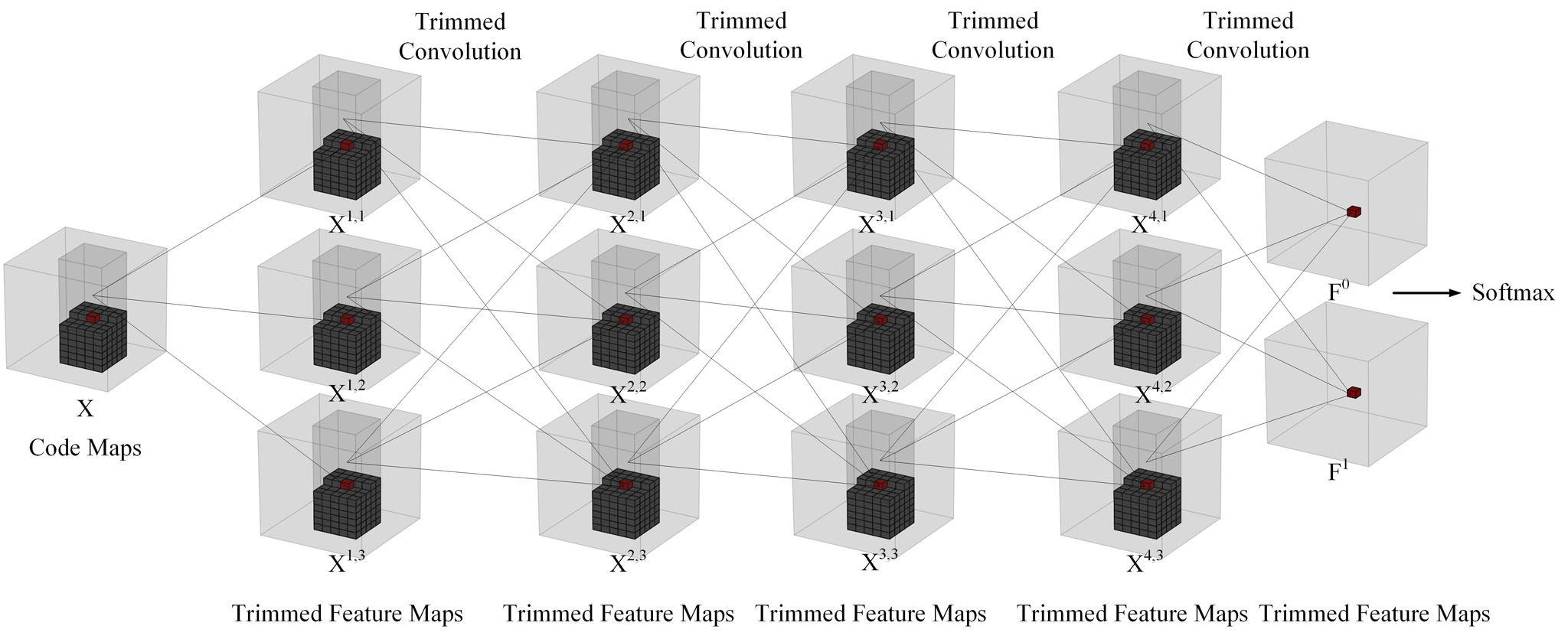}%
\subcaption{Framework for trimmed convolutional network for entropy probability estimator. The whole code maps are processed as a sample.}\label{fig:1b}
\end{minipage}%
\caption{Comparison between trimmed convolutional networks and traditional convolutional networks for entropy probability estimator.} \label{fig2}
\end{figure}

\section{Trimmed Convolutional Arithmetic Encoding}\label{sec:method}

In this section, we present our trimmed convolutional arithmetic encoding model (TCAE).
As illustrated in Fig.~\ref{fig:1b}, TCAE utilizes the trimmed convolutional networks to predict the probability of codes from their context in one single forward pass.
Before introducing the trimmed convolutional network, we first describe arithmetic coding and coding context in Sections~\ref{s3_1} and~\ref{s3_2}.
Then trimmed convolutional network is provided for context modeling and probability prediction in Section~\ref{s3_3}, and a variant of TCAE, the slope TCAE is proposed for parallel decoding.  The model objective is described in Section~\ref{s3_4}.

\subsection{Arithmetic coding}\label{s3_1}

Arithmetic encoding is an entropy encoding scheme for lossless data compression.
Given a string of symbols, entropy encoding assigns fewer bits for frequently occurring symbols and more bits for not-so-frequently occurring ones.
Different from other entropy encoding algorithms such as Huffman coding, arithmetic coding encodes the entire string of symbols into a single number in the interval of $[0,1]$.
Denote by $k$ the number of symbols occurred in the coding system.
Given a new symbol $x_i$ and the current interval, arithmetic coding first predicts the probabilities $p(x_i=t|x_{i-1},\ldots,x_0)$ that the new symbol belongs to each value $t$, and the current interval is further divided according to the predicted probabilities.
In order to encode the updated sequence, the current interval is then updated based on the ground truth of the new symbol as well as its predicted probability.
For example, Fig.~\ref{arithmetic} illustrates the arithmetic coding result of a sequence (0, 2, 3) for a coding system with the symbols (0, 1, 2, 3) and the discrete distribution with probabilities (0.6, 0.2, 0.1, 0.1).



\subsection{Coding schedule and context of 3D cuboid}\label{s3_2}
\begin{figure}[!tbp]
\centering
\begin{minipage}[b]{0.45\textwidth}
  \centering
\includegraphics[width=\textwidth]{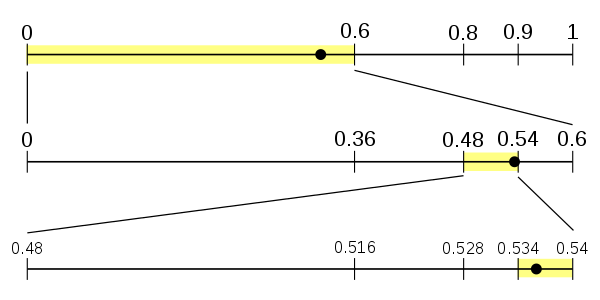}%
\subcaption{Arithmetic encoding of a sequence (0,2,3) for a coding system with the symbols (0,1,2,3) and the i.i.d distribution with probabilities (0.6,0.2,0.1,0.1)}\label{arithmetic}
\end{minipage}
\hfill
\begin{minipage}[b]{0.45\textwidth}
  \centering
\includegraphics[width=\textwidth]{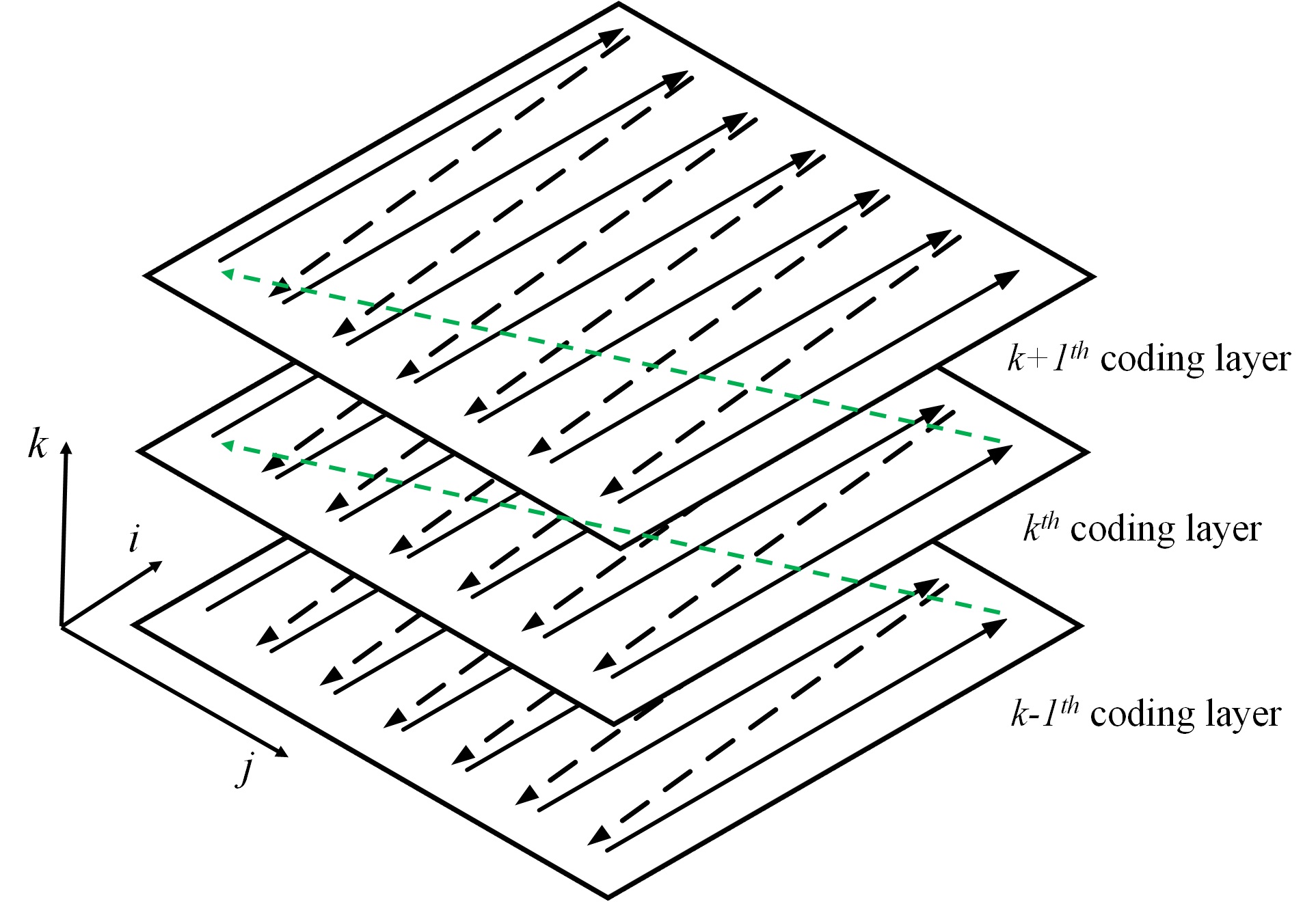}%
\subcaption{Coding schedule for binary code cuboid.}\label{schedule}
\end{minipage}%
\caption{Arithmetic encoding and coding schedule.} 
\end{figure}

In this work, we focus on the arithmetic encoding of 3D binary code cuboid $\mathbf{X} = \{x_{i,j,k}| 0 \leq i \leq W-1, 0 \leq j \leq H-1, 0 \leq k \leq C-1\}$.
As illustrated in Fig.~\ref{schedule}, beginning at $x_{0,0,0}$, we follow the following order to encode $\mathbf{X}$:
(i) $x_{i+1,j,k}$ is encoded after $x_{i,j,k}$ until $i = W-1$;
(ii) when $i = W-1$, $x_{0,j+1,k}$ is encoded after $x_{i,j,k}$ until $j = H-1$;
(iii) when $i = W-1$ and $j = H-1$, $x_{0,0,k+1}$ is encoded after $x_{i,j,k}$.

One key issue of arithmetic encoding is to predict the probability of a symbol from its context, i.e., the preceding encoded symbols.
Given the position $(p,q,r)$ of 3D cuboid, we denote its full context as $CTX(x_{p,q,r})$.
Taking the encoding order of $\mathbf{X}$ into account, $CTX(x_{p,q,r})$ can be defined as $CTX(x_{p,q,r}) = \{x_{i,j,k}| \{k<r\} \vee \{ k=r, j<q \} \vee \{k=r,j=q,i<p\}\}$.
Unfortunately, the length of the full context $CTX(x_{p,q,r})$ is not fixed and varies by the position $(p,q,r)$, making it difficult to learn probability prediction model on $CTX(x_{p,q,r})$.
Note that the context close to the bit to be encoded plays more important role in probability prediction.
Thus we adopt a fixed length context defined as $CTX_f(x_{p,q,r}) = \{x_{i,j,k}| \{ r-c \leq k < r, |i-p| \leq w, |j-q| \leq h\} \vee \{ k=r, q-h \leq j < q, |i-p| \leq w \} \vee \{k=r,j=q, p-w \leq i < p\}\}$.
It is natural to expect that large context (i.e., large $w, h, c$) benefits probability prediction, and in the following we suggest to use trimmed convolutional networks for improved and efficient arithmetic encoding.

\subsection{Trimmed convolutional networks}\label{s3_3}
With the fixed length context defined above, we can extract a $(2w+1) \times (2h+1) \times (c+1)$ cuboid $\mathbf{x}_{p,q,r} = \{x_{i,j,k} | \{ r-c \leq k \leq r, p-w \leq i \leq p+w, q-h \leq j \leq q+h \}$.
As shown in Fig.~\ref{fig:1a}, the blue voxels represent the encoded bits before ${x}_{p,q,r}$, the red voxel denotes the next bit to be encoded, and the gray ones are the other non-encoded bits.
Then, for each location ${x}_{p,q,r}$, Li et al.~\cite{li2017learning} adopt a CNN architecture of three convolution layers followed by three fully connected layers to predict ${x}_{p,q,r}$ from its context cuboid $\mathbf{x}_{p,q,r}$.

One main obstacle, which restricts the shared computation of the probability prediction, is that the introduction of non-encoded bits (i.e., the red and gray voxels in Fig.~\ref{fig:1a}).
In~\cite{li2017learning}, Li et al. suggest to assign a special default value to the current non-encoded bits in the context cuboid $\mathbf{x}_{p,q,r}$.
For example, when predicting ${x}_{p,q,r}$, ${x}_{p,q,r}$ should be replaced with default value~\cite{li2017learning}.
When encoding the next bit after ${x}_{p,q,r}$, the original value of ${x}_{p,q,r}$ should be used in context modeling.
Therefore, fully convolutional network cannot be directly utilized for context modeling, and the probability prediction is performed independently for each ${x}_{p,q,r}$, leading to repeating computation and encoding inefficiency~\cite{li2017learning}.

Fortunately, the value of each location only have two choices, i.e., the original value ${x}_{p,q,r}$ and the default value for non-encoded bit.
Moreover, given the voxel ${x}_{p,q,r}$ to encode, the positions of all non-encoded bits are also fixed.
Without loss of generality, we let the default value be 0.
Thus, we can introduce a group of trimmed convolution operators to incorporate with fully convolutional network for context modeling.
Denote by $\mathbf{w}^0$ a 3D convolution kernel $\mathbf{w}^0 = \{ w_{t,i,j,k}| -w_0 \leq i \leq w_0, -h_0 \leq j \leq h_0, -t \leq k \leq C-t-1\}$.
Then the convolution result $(\mathbf{X}*\mathbf{w}^0)$ at the location $(p,q,r)$ can be written as,
\begin{equation}
\label{eqn:convolution}
(\mathbf{X}*\mathbf{w}^0)(p,q,r) = \sum_{l-i = p, m-j = q, n-k = r,t=r} x_{l,m,n}w^0_{t,i,j,k}.
\end{equation}
However, both the context and non-encoded bits have effect on the convolution result $(\mathbf{X}*\mathbf{w}^0)(p,q,r)$.
We note that the positions of non-encoded bits are fixed and pre-defined with respect to $\mathbf{w}^0$, and thus can introduce a mask $\mathbf{m}$ of $\{0, 1\}$ to exclude them in convolution.
To this end, $m_{i,j,k}$ is defined as $1$ if $x_{p+i,q+j,r+k}$ is encoded before $x_{p,q,r}$, and 0 otherwise.
The trimmed convolution is thus defined as,
\begin{equation}
\label{eqn:trimmed_convolution}
\mathbf{X}^{1} = \mathbf{X}*(\mathbf{m} \circ \mathbf{w}^0),
\end{equation}
where $*$ denotes the convolution operator, and $\circ$ denotes the element-wise multiplication operator.
With trimmed convolution, we can safely avoid the effect of non-encoded bits in context modeling while maintaining the efficiency of fully convolutional network for predicting probabilities of all bits in one forward pass.

In the following, we first use single convolution kernel as an example to explain the settings of $\mathbf{m}$, 
%
%
which are different for the input layer and the hidden layers.
For the input layer, when predicting the probability of $x_{p,q,r}$, both $x_{p,q,r}$ and the bits encoded after $x_{p,q,r}$ should be masked out in trimmed convolution.
Following the definition of context in Section~\ref{s3_2}, we define the mask $\mathbf{m}^0$ for the input layer as,
\begin{equation}
{m}^0_{ijk}=\begin{cases}
1,  \mbox{if} (k \!<\! 0) \!\vee\! (k \!=\! 0, j \!<\! 0) \!\vee\! (k \!=\! 0, j \!=\! 0, i \!<\! 0)  \\
0, \mbox{otherwise}
\end{cases}
\end{equation}
When it comes to the hidden layer $\mathbf{X}^{d}$ ($d \geq 1$), we note that $x^{d}_{p,q,r}$ only conveys the context information of $x_{p,q,r}$ and should not be excluded in the further context modeling.
Therefore, we modify the definition of the mask $\mathbf{m}^d$ ($d \geq 1$) for hidden layer as,
\begin{equation}
{m}^d_{ijk}=\begin{cases}
1,  \mbox{if} (k \!<\! 0) \!\vee\! (k \!=\! 0, j \!<\! 0) \!\vee\! (k \!=\! 0, j \!=\! 0, i \!\leq\! 0)  \\
0, \mbox{otherwise}.
\end{cases}
\end{equation}

\begin{figure}[!tbp]
\begin{center}
\includegraphics[width=0.9\linewidth]{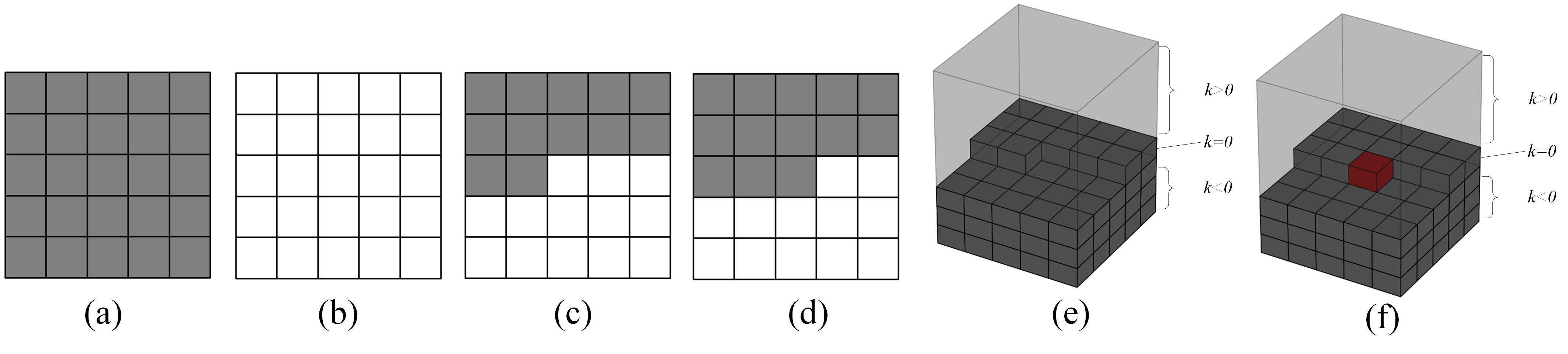}
\end{center}
   \caption{Mask planes with respect to $k$ for trimmed convolution kernels with the size of $5\times5$. The gray value and red value denotes 1 and the white value denotes 0.
    (a) $k<0$, (b) $k>0$, (c) $k = 0$ for the input layer, (d) $k = 0$ for the hidden layers, (e) 3D kernel mask for input layer, (f) 3D kernel mask for hidden layer.
}\label{mask}
\end{figure}

Using a 3D convolution kernel with the size of $5 \times 5 \times C$ as an example, Fig.~\ref{mask} illustrates the representative mask planes with respect to $k$.
As shown in Fig.~\ref{mask}ab, when $k < 0$ ($k > 0$), the $k$-th mask plane is a matrix of 1s (0s) for both the input layer and the hidden layers.
when $k = 0$, the center position is masked out in the mask plane for the input layer (see Fig.~\ref{mask}c) but are included for the hidden layers (see Fig.~\ref{mask}d).

The trimmed convolution in Eqn. (\ref{eqn:trimmed_convolution}) only uses one set of convolution kernels in each layer, which is limited in probability prediction.
Thus, we extend the trimmed convolution to the multiple convolution kernel form.
Suppose there are $g_{in}$ groups of feature maps $\mathcal{X}^d = \{\mathbf{X}^{d,1}, ...,  \mathbf{X}^{d,g_{in}}\}$ in the $d$-th layer and $g_{out}$ groups of feature maps $\mathcal{X}^{d+1} = \{\mathbf{X}^{d+1,1}, ...,  \mathbf{X}^{d+1,g_{out}}\}$ in the $(d+1)$-th layer.
Each group of feature map has the same size with the input  cuboid $\mathbf{X}$.
The group trimmed convolution is then defined as,
\begin{equation}
\label{eqn:group_trim}
\mathbf{X}^{d+1, g^{\prime}} = \sum_{g = 1}^{g_{in}} \mathbf{X}^{d,g}*(\mathbf{m}^d \circ \mathbf{w}^{d,g,g^{\prime}}),
\end{equation}
where $\mathbf{X}^{d,g}$ denotes the $g$-th group of feature map in $\mathcal{X}^d$, $\mathbf{X}^{d+1, g^{\prime}}$ denotes the $g^{\prime}$-th group of feature map in $\mathcal{X}^{d+1}$, $\mathbf{m}^d$ is the mask for the $d$-th layer, and $\mathbf{w}^{d,g,g^{\prime}}$ is the convolution kernel to connect $\mathbf{X}^{d,g}$ and $\mathbf{X}^{d+1, g^{\prime}}$.

\subsection{Slope TCAE}\label{s3_4}
With the proposed TCAE, the possibility of all the codes can be predicted simultaneously with the trimmed convolutional network in coding period. However, in the decoding stage, the prediction of the possibility of $x_i$, $p(x_i|x_{i-1},\ldots,x_0)$, depends on $x_{i-1},\ldots,x_0$, making that $x_i$ should be decoded after $x_{i-1},\ldots,x_0$. 
Thus, the possibility prediction of codes still need to be processed in serial order. This becomes a bottleneck of the deep learning based context modeler which is usually speed up with GPU and parallel computation. 

To speed up the decoding process of TCAE, we should break down some dependency among codes. That is to say, for a code $x_i$, it does not depend on all the $x_{i-1},\ldots,x_0$ but only part of them. By dividing the codes $\mathbf{x}$ into different blocks and supposing the codes inner block are independent, we can break down the dependency of the codes inner block and parallel predict the codes in one block simultaneously. Here, the $t$-\@th code block is defined as:
\begin{equation}
CB_t(\mathbf{x})=\{x_{i,j,k}|\;i+j+k=t\}.
\end{equation}
Then, the possibility of $x_i$ can be modeled as $p(x_i|\;CB_{t-1}(\mathbf{x}),\ldots,CB_0(\mathbf{x}))$ where $x_i \in CB_t(\mathbf{x})$. Thus, all the $x_i$ in the code block $CB_t(\mathbf{x})$ share the same context and can be predicted in the same time. Since the codes in one block exactly fall on a slope plane of a cuboid, we call the parallel context modeler as slope TCAE. 

With the slope TCAE, the context should be modified. The context for code $x_{p,q,r}$, $CTX(x_{p,q,r})=\{x_{i,j,k}|\;i+j+k<p+q+r\}$. For feature context, all the features in the plane $i+j+k=p+q+r$ are predicted with the context $CTX(x_{p,q,r})$ and can be further used as the context of the feature $f_{p,q,r}$ to predict $x_{p,q,r}$, $CTX_f(f_{p,q,r})=\{f_{i,j,k}|\;i+j+k\leq p+q+r\}$.

The coding schedule is refined as coding the 3D codes map according to the defined blocks, from $CB_0(\mathbf{x})$ to $CB_{H+W+C}(\mathbf{x})$. Inner one block $CB_t(\mathbf{x})$, we sort the elements in ascending order first with the index $k$ then with the index $i$. The new schedule is used for slope TCAE.
Further more, the mask $\mathbf{m}^0$ and $\mathbf{m}^d$ are modified as:
\begin{equation}
{m}^0_{ijk}=\begin{cases}
1,  \mbox{if} \;\; i+j+k<0  \\
0, \mbox{otherwise}
\end{cases}
\end{equation}
\begin{equation}
{m}^d_{ijk}=\begin{cases}
1,  \mbox{if} \;\; i+j+k\leq 0  \\
0, \mbox{otherwise}.
\end{cases}
\end{equation}

\subsection{Model objective and learning}\label{s3_5}
Given all the model parameters $\mathcal{W} = \{ \mathbf{w}^{d,g,g^{\prime}} \}$ , the output of the trimmed convolutional network can be written as $F(\mathbf{X}; \mathcal{W})$.
$F(\mathbf{X}; \mathcal{W})$ includes $m$ parts, $\left(F(\mathbf{X}; \mathcal{W})\right)^t_{p,q,v}$ denotes the predicted probability of $x_{p,q,v} = t$ with $t=0,\ldots,m-1$. And $m$ is the number of different code value in the input code maps.
We adopt the length of the codes after arithmetic encoding as the model objective,
\begin{equation}
\label{eqn:objective}
\ell(\mathcal{W}; \mathbf{X}) = \sum_{{p,q,v}}\sum_{t=0}^{m-1} -s(x_{p,q,v},t) \log_2 \left(F(\mathbf{X}; \mathcal{W})\right)^t_{p,q,v}
\end{equation}
where $s(x_{p,q,v},t)=1$ when $x_{p,q,v} = t$, and $s(x_{p,q,v},t)=0$ otherwise.
According to the Shannon's theorem, the compression ratio is defined as the ratio between uncompressed size and compressed size, and can be written as
\begin{eqnarray}
\frac{1}{r(e)}&=&\frac{\ell(\mathcal{W}; \mathbf{X})}{CHW \log_2 m}=\frac{1}{\log_2 m}[\frac{1}{CHW}\nonumber\\
\!\!&\!\!\!\!&\!\! \sum_{{p,q,v}}\sum_{t=0}^{m-1} -s(x_{p,q,v},t) \log_2 \left(F(\mathbf{X}; \mathcal{W})\right)^t_{p,q,v}]. 
\end{eqnarray}
Thus, it is reasonable to use the objective in Eqn. (\ref{eqn:objective}) to learn probability prediction model for entropy encoding.

By minimizing the model objective defined in Eqn. (\ref{eqn:objective}), the trimmed convolutional network $F(\mathbf{X}; \mathcal{W})$ can be learned from training data in an end-to-end manner.
In this work, we adopt the ADAM solver~\cite{kingma2014adam} to learn the model parameters.
The model is trained with the learning rate of $3 \times 10^{-4}$, $1 \times 10^{-4}$, $3.33 \times 10^{-5}$ and $1.11 \times 10^{-5}$.	
The smaller learning rate is adopted until the objective with the larger one stops decreasing.

\section{Experiments}\label{sec:result}
Three groups of experiments are conducted to test the proposed trimmed convolutional arithmetic encoding (TCAE) and slope TCAE.
The first is the lossless compression of gray image.
To satisfy the requirement of 3D binary cuboid, we take the 8-bit representation of gray image as the input. 
To verify the context modeling ability of TCAE, we further used the trained gray image predictor for image inpainting.
%
The third group of experiments is the incorporation with CNN-based lossy compression.
We use TCAE to compress the binary codes and importance map generated by~\cite{li2017learning}, and compare the rate-distortion performance with the baseline method~\cite{li2017learning}, JPEG, JPEG-2000, and Ball\'{e} et al.~\cite{balle2016end}.
%
\begin{figure}[!tbp]
\begin{center}
\includegraphics[width=1.0\linewidth,height=0.26\textheight]{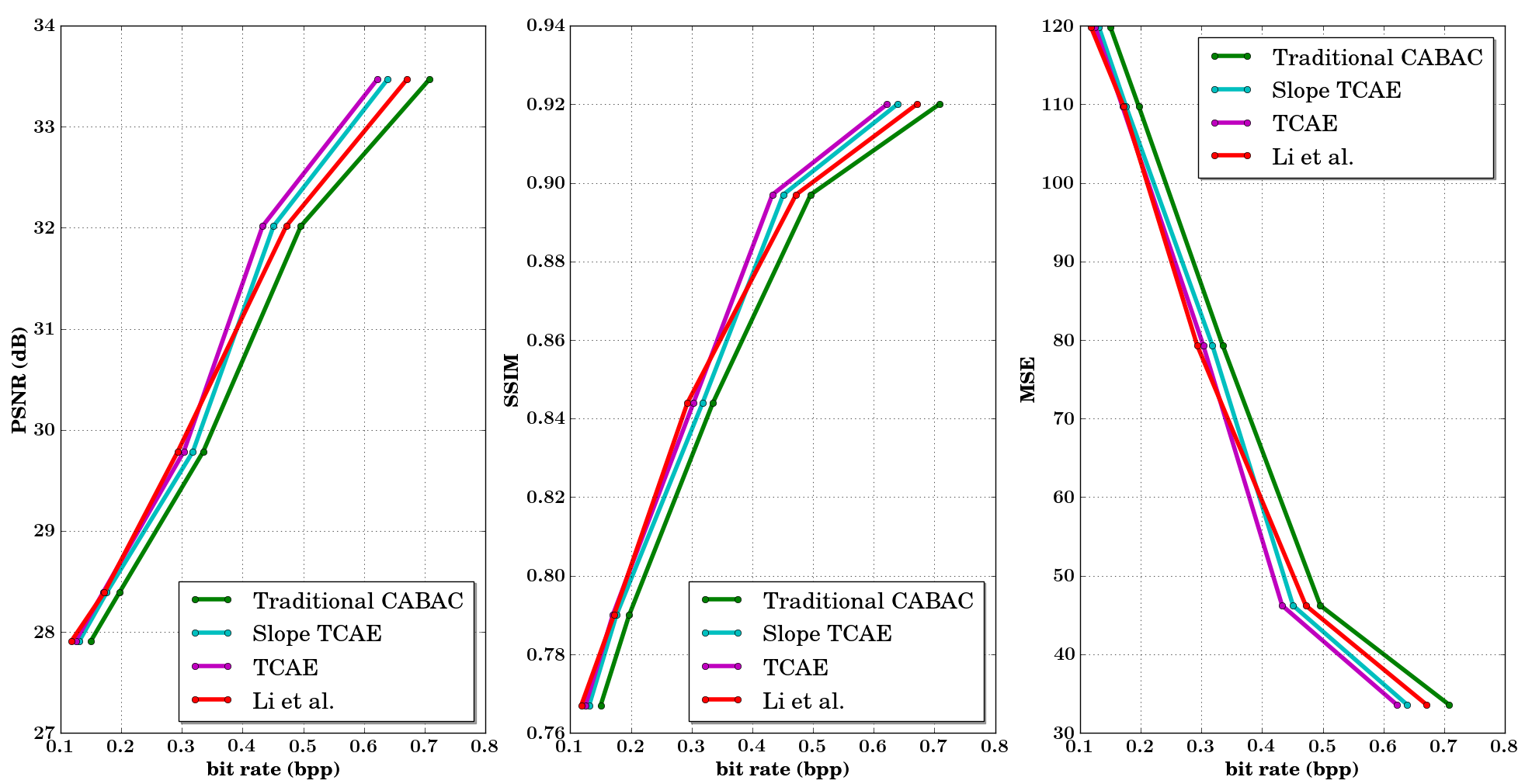}
\end{center}
   \caption{The effect of TCAE and slope TCAE on the rate-distortion performance of the lossy image compression system~\cite{li2017learning}}\label{bpp}
\end{figure}
\subsection{Network architecture and parameter setting}

\begin{table}[!tb]
\scriptsize
\begin{center}
\caption{Architecture of trimmed convolutional networks.}\label{table:predictor}
\begin{tabular}{c|c}
\hline
Layer & Activation size \\
\hline
Input & $h\times w \times c$ \\
$g\times c\times 5\times 5 \times c$ trimmed conv, pad $2$, stride $1$ & $g\times h\times w\times c$ \\
$g\times c\times 5\times 5 \times c$ trimmed conv, pad $2$, stride $1$ & $g\times h\times w\times c$ \\
Trimmed residual block, $g\times c$ 3D filters & $g\times h\times w\times c$ \\
Trimmed residual block,  $g\times c$ 3D filters & $g\times h\times w\times c$ \\
Trimmed residual block,  $g\times c$ 3D filters & $g\times h\times w\times c$ \\
Trimmed residual block,  $g\times c$ 3D filters & $g\times h\times w\times c$ \\
$m\times c \times 5\times 5\times c$ trimmed conv, pad $2$, stride $1$ & $m\times h\times w\times c$ \\
\hline
\end{tabular}
\end{center}
\end{table}
Table~\ref{table:predictor} gives a general network structure for the TCAC. Let the size of the input 3D code map is $w \times h \times c$. $g$ is the number of groups used in each convolution layer. $m$ is the number of labels to be predict. The output represents the possibility the code belongs to each class i.e. $\left(F(\mathbf{X}; \mathcal{W})\right)^t_{p,q,v}$ . $p,q,v$ is the index of the code in the 3D code map $\mathbf{X}$ and $t$ is the index of each class with $t=0,\ldots,m-1$.
In TCAE, we also introduce the trimmed residual block, which consists of two trimmed convolution layers with each followed by the ReLU nonlinearity.
The skip connection is also added from the input to the output of the residual block.
For defining the model objective in Eqn~\ref{eqn:objective}, feature map rearrangement is adopted to reshape the input and output of the trimmed convolutional network.
Concretely, the input is reshaped into a one dimensional vector with the size of $chw \times 1$, while the output is reshaped into a 2D matrix with the size of $chw \times m$.

All the networks are trained on $10,000$ high quality images from the ImageNet~\cite{deng2009imagenet} and tested on the Kodak PhotoCD image dataset.
Due to that TCAE is a fully convolutional network, it can be trained and tested using images with any size.
In the experiments, we set the image size as $128 \times 128$ for training lossless gray image compression modeler.
Note that the input of TCAE is required to be 3D binary cuboid.
Thus, we transform the discrete gray image to its 8-bit representation, i.e. 8 binary bit planes, resulting in a $128 \times 128 \times 8$ 3D binary cuboid.
\subsection{Lossless gray image compression}
For lossless gray image compression, our TCAE is simply deployed to the 8-bit representation of gray image without any transform.
We compare TCAE with several popular lossless compression standards, including GIF, TIFF, PNG, JPEG-LS and JPEG2000-LS.
For PNG, JPEG2000-LS, JPEG-LS, GIF and TIFF, all the compression ratio is calculated with the results generated with the Matlab2015b.
The average compression ratio on 24 image from Kodak set are given in Table~\ref{table:lossless}.
Our TCAE achieves the best compression ratio (2.00), which is much better than the second best method (1.79 for JPEG2000-LS).

\begin{table}[htb]
\scriptsize
\begin{center}
\caption{Results for lossless gray image compression.}
\begin{tabular}{|c|c|c|c|c|c|c|c|}
\hline
Method & TIFF&GIF&PNG&JPEG-LS&JPEG2000-LS&TCAE&slope TCAE\\
\hline
Compression Ratio&$1.01$&$1.13$&$1.61$&$1.57$&$1.79$&${\color{red}2.00}$&$1.99$\\
\hline
\end{tabular}
\label{table:lossless}
\end{center}
\end{table}

\subsection{Image inpainting}
\begin{figure}[!tbp]
\begin{center}
\includegraphics[width=1.0\linewidth]{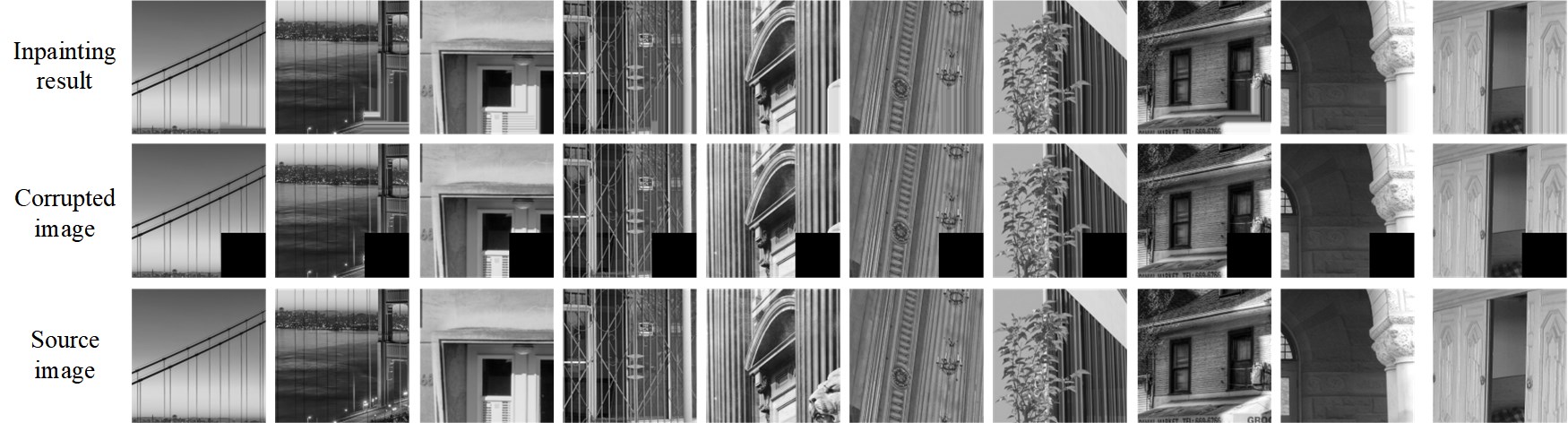}
\end{center}
   \caption{Inpainting results of TCAE.}\label{inpaint}
\end{figure}
With the trained model on lossless image compression, we apply it for the job of image inpainting. We corrupt a rectangle area with the size of $\frac{1}{9}$ from the right bottom of the original image and then fill in the blank area with the TCAE context predictor. We just assign $0$ or $1$ to each binary plan of the gray image in the blank area with the probability predicted by the TCAE. Figure~\ref{inpaint} gives the inpainting results of TCAE, which shows the TCAE has strong ability in recovering the edges.
\subsection{CNN-based lossy image compression~\cite{li2017learning}}
\begin{figure}[!tbp]
\begin{center}
\includegraphics[width=1.0\linewidth,height=0.26\textheight]{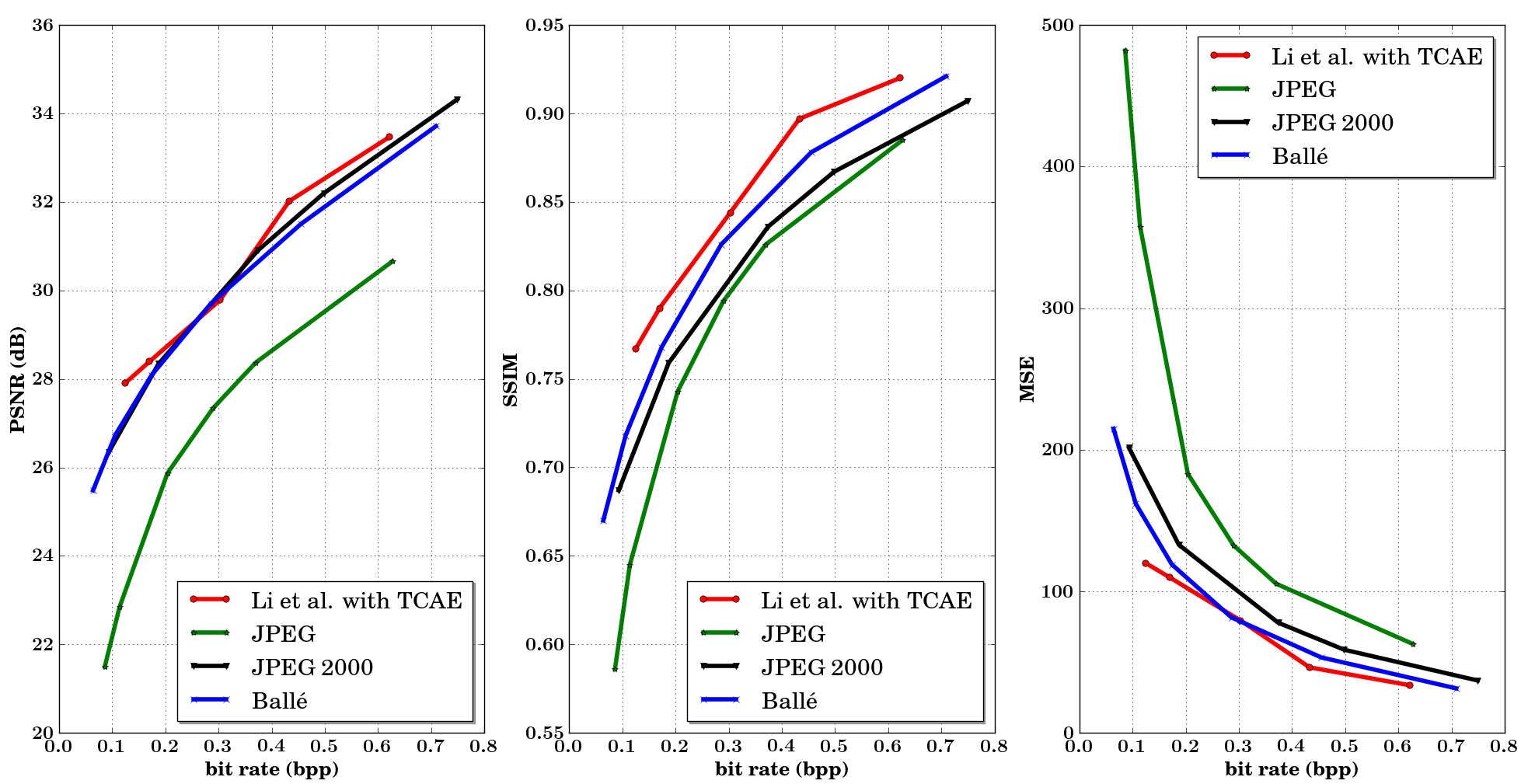}
\end{center}
   \caption{Comparision of  rate-distortion performance of the TCAE extention of~\cite{li2017learning} and other lossy image compression methods.}\label{cmp}
\end{figure}

Entropy encoding can also be used in the CNN-based lossy image compression system to compress the intermediate codes.
In our experiment, we take the system~\cite{li2017learning} as an example, and replace the convolutional entropy encoder in~\cite{li2017learning} with our TCAE.
Fig.~\ref{bpp} shows the rate-distortion curves obtained using our TCAE and convolutional entropy encoder~\cite{li2017learning}.
Here, SSIM, MSE, and PSNR are adopted as the distortion performance metrics, and bit per pixel ($bpp$) is used as the indicator of compression rate.
As shown in Fig.~\ref{bpp}, our TCAE and slope TCAE achieves comparable performance with convolutional entropy encoder~\cite{li2017learning}. And all the convolutional network based context models, e.g., TCAE, slope TCAE and the convolutional entropy encoder, can get better compression ratio than the traditional CABAC context modeler. 
When $bpp \geq 0.4$, TCAE and slope TCAE performs better than convolutional entropy encoder~\cite{li2017learning}, which can be attributed to the ability of TCAE in modeling large context.
Fig.~\ref{cmp} further compares the rate-distortion curves obtained using Li et al.~\cite{li2017learning} with TCAE, JPEG, JPEG 2000, and Ball\'{e}~\cite{balle2016end}.
When $bpp \geq 0.4$, Li et al.~\cite{li2017learning} with TCAE outperforms the competing methods, regardless of any distortion metrics.
The result indicates that TCAE can be incorporated with the CNN-based lossy image compression to boost rate-distortion performance.

The most prominent merit of TCAE is its efficiency in encoding due to the incorporation of trimmed convolution and fully convolutional network.
To illustrate this, Table~\ref{table:time} reports the run time for encoding the intermediate codes~\cite{li2017learning}.
Here, Groups 1$\sim$5 represent the five settings of model parameters~\cite{li2017learning} to obtain the compression models at different $bpp$s in the range of $[0.118, 0.671]$.
All the experiments are conducted on a computer with the GTX TitanX GPU of 12GB memory.
The reported time is based on the encoding of the importance map and binary codes of a $752\times496\times3$ image, and we do not include the time for generating the intermediate codes.
In general, it takes about 0.051 $s$ and 0.098 $s$ to generate the intermediate codes with 64 and 128 channels, respectively.
From Table~\ref{table:time}, it can be seen that our TCAE is 7$\sim$20 times faster than the entropy encoder used in~\cite{li2017learning}.
With the introduction of trimmed convolution, our TCAE can be performed at a near real time speed (e.g., 25 fps ) to encode the intermediate codes of images with the size of $752\times496\times3$.
\subsection{Slope TCAE vs. TCAE}
As shown in Table~\ref{table:lossless} and Figure~\ref{bpp}, slope TCAE only has small drop both in gray image compression and compressing 3D code maps due to removing the dependency among codes inner code blocks.
It verifies our assumption, the codes inner one block are independent, is reasonable.
For the encoding efficiency, the encoding speed of slope TCAE is the same as TCAE. 
For the decoding efficiency, with the small drop of the compression ratio, the slope TCAE get a huge improvement in decoding efficiency. 

In decoding period, the convolutional entropy encoder~\cite{li2017learning} uses about $212.32$ seconds to decompress the 3D binary code map with the size of $94\times 68\times 64$. Cropping the 3D binary codes into smaller size can speed up the decoding speed. With experiments, by cropping the 3D maps into $16\times16\times64$, the compression ratio has no clear drop. With the strategy, it takes $15.64$ seconds to decode one image for convolutional entropy encoder~\cite{li2017learning} by parallel decoding several small 3D code maps. With the same setting, TCAE takes $42.31$ seconds to decode the 3D code blocks. However, the slope TCAE only needs $0.62$ seconds for decoding the 3D code maps, which is more than $60\times$ faster than TCAE and $25$ times faster than convolutional entropy encoder~\cite{li2017learning}.

\begin{table}[!tb]
\scriptsize
\begin{center}
\caption{Run time (in seconds, $s$) used for encoding the intermediate codes in~\cite{li2017learning}.}
\begin{tabular}{c|c|c|c|c|c|c}
\hline
Group&\multicolumn{3}{|c|}{Convolutional entropy encoder~\cite{li2017learning} }& \multicolumn{3}{|c}{TCAE} \\
\hline
&Binary&Importance&Total&Binary&Importance&Total\\
\hline
Group 1 & $0.181$&$0.092$&$0.274$&$0.035$&$0.003$&$0.038$\\
\hline
Group 2 & $0.249$&$0.092$&$0.342$&$0.035$&$0.003$&$0.038$\\
\hline
Group 3 & $0.468$&$0.092$&$0.561$&$0.035$&$0.003$&$0.038$\\
\hline
Group 4 & $0.710$&$0.092$&$0.802$&$0.035$&$0.003$&$0.038$\\
\hline
Group 5 & $1.199$&$0.115$&$1.313$&$0.161$&$0.005$&$0.166$\\
\hline
\end{tabular}
\label{table:time}
\end{center}
\end{table}
%
\section{Conclusion}\label{sec:conclusion}
The paper presents a trimmed convolutional network for arithmetic encoding (i.e., TCAE) of 3D binary cuboid.
Benefitted from trimmed convolution, we can utilize the fully convolutional network for performing probability prediction to all bits in one single forward pass, making it feasible to model large context while maintaining computational efficiency.
To speed up the decoding process, we introduce a slope TCAE schedule to divide the 3D code maps into different blocks and remove dependency among codes inner blocks for parallel decoding.
In comparison with traditional methods such as PNG and JPEG2000-LS, our TCAE and slope TCAE achieve better compression ratio for lossless gray image compression.
Moreover, it can also be incorporated with the CNN-based lossy image compression system (e.g.,~\cite{li2017learning}) to compress the intermediate codes, and exhibits its superiority in large context modeling and real time encoding speed.
%
%

\clearpage

\bibliographystyle{splncs}
\bibliography{egbib}
\end{document}